\documentclass{bmvc2k}
\usepackage{multirow}
\usepackage{amsfonts}
\usepackage{stackengine}
\usepackage{adjustbox}

\title{BI-GCN: Boundary-Aware Input-Dependent \\ Graph Convolution Network for \\ Biomedical Image Segmentation}

\addauthor{Yanda Meng}{}{1}
\addauthor{Hongrun Zhang}{}{1}
\addauthor{Dongxu Gao}{}{2}
\addauthor{Yitian Zhao}{}{3}
\addauthor{Xiaoyun Yang}{}{4}
\addauthor{Xuesheng Qian}{}{5}
\addauthor{Xiaowei Huang}{}{6}
\addauthor{Yalin Zheng*}{yalin.zheng@liverpool.ac.uk}{1}

\addinstitution{
 Department of Eye and Vision Science, University of Liverpool, Liverpool, UK.
}
\addinstitution{
 School of Computing, University of Portsmouth, Portsmouth, UK.
}
\addinstitution{
 Cixi Institute of Biomedical Engineering, Chinese Academy of Sciences, Ningbo, China.
}
\addinstitution{
 Remark AI UK Limited, London, United Kingdom.
}
\addinstitution{
 China Science IntelliCloud Technology Co., Ltd, Shanghai, China.
}
\addinstitution{
 Department of Computer Science, University of Liverpool, Liverpool, UK.
}

\runninghead{Y. Meng et al.}{Boundary-Aware Input-Dependent GCN}

\begin{document}

\maketitle

\begin{abstract}
Segmentation is an essential operation of image processing. The convolution operation suffers from a limited receptive field, while global modelling is fundamental to segmentation tasks. In this paper, we apply graph convolution into the segmentation task and propose an improved \textit{Laplacian}. Different from existing methods, our \textit{Laplacian} is data-dependent, and we introduce two attention diagonal matrices to learn a better vertex relationship.
In addition, it takes advantage of both region and boundary information when performing graph-based information propagation. 
Specifically, we model and reason about the boundary-aware region-wise correlations of different classes through learning graph representations, which is capable of manipulating long range semantic reasoning across various regions with the spatial enhancement along the object's boundary.
Our model is well-suited to obtain global semantic region information while also accommodates local spatial boundary characteristics simultaneously. 
Experiments on two types of challenging datasets demonstrate that our method outperforms the state-of-the-art approaches on the segmentation of polyps in colonoscopy images and of the optic disc and optic cup in colour fundus images.
\end{abstract}

\section{Introduction}
Accurate examination of anatomical structures in biomedical images helps clinicians determine the likelihood of many medical conditions and diseases.
For example, shapes of the optic disc (OD) and optic cup (OC) of the retina are essential in the diagnosis of glaucoma, a chronic degenerative disease that causes irreversible loss of vision \cite{tham2014global}. 
Likewise, colorectal polyps are linked with an increased risk of having colorectal cancer, the third most common cancer worldwide \cite{silva2014toward}. 
However, it is impractical to manually annotate these structures in clinics due to the time-consuming, labour-intensive, and error-prone process. Automated and accurate image segmentation is increasingly needed. 

Deep learning-based pixel-wise biomedical image segmentation methods, such as \cite{ronneberger2015u,zhou2018unet++,fu2018joint,fan2020pranet,wu2019oval,zhang2020adaptive}, are popular because of Convolutional Neural Network's (\textit{CNN's}) excellent ability to extract and reason about semantic features, and correlations among different classes.
\textit{Fan et al.} \cite{fan2020pranet} and \textit{Zhang et al.} \cite{zhang2020adaptive}
studied the correlations between the foreground and background classes using a similar foreground erasing mechanism, which achieved state-of-the-art performance in the colonoscopy polyps segmentation tasks. Due to limited receptive fields of \textit{CNN}, they incorporated dense dilation \cite{YuKoltun2016} convolutions to enlarge the receptive regions for long range context reasoning. Along the same line, \textit{Fu et al.} \cite{fu2018joint} proposed multiscale input and side-output mechanism with deep supervision, which achieves multi-level receptive field fusion for long range relationship aggregation.
However, this is inefficient since stacking local cues cannot always precisely handle long range context relationships. Especially for pixel-level classification problems, such as segmentation, performing long range interactions is an important factor for reasoning in complex scenarios \cite{chen2017deeplab}. For example, it is prone to assign visually similar pixels in a local region into the same category. Meanwhile, pixels of the same object but distributed in a distance are difficult to construct dependencies.
Recent graph propagation-based reasoning approaches \cite{liang2018symbolic,chen2019graph,zhangli_dgcn,ke2021deep} that use \textit{Laplacian} smoothing-based graph convolution \cite{kipf2016semi}, provide specific benefits in the sense of global long range information reasoning.
They estimated the initial graph structure from a data-independent \textit{Laplacian} matrix defined by randomly initialized adjacency matrix \cite{chen2019graph,zhangli_dgcn} or hand-crafted adjacency matrix \cite{liang2018symbolic,kipf2016semi}. However, one may make a model to learn a specific long range context pattern \cite{li2020spatial}, which is less related to the input features. Differently, as seen in previous works, the graph structure can be estimated with the similarity matrix from the input data \cite{li2018beyond}, we estimate the initial adjacency matrix in a data-dependent way. Specifically, we implement two diagonal matrices to perform channel-wise attention on the dot product distance of input vertex embeddings, and measure spatial-wise weighted relations among different vertices. In this way, our model is capable of learning an input-related long range context pattern, which improves the model segmentation performance, please read \textit{Ablation Study} (Section~\ref{sec:ablation}) for more details.

Different to the dense pixel classification, 
recent polygon-based methods \cite{meng2020regression,meng2020cnn} explored the spatial features of an object's boundary. 
They regressed the predefined vertex position along the object borders then linked the predicted vertices to form a polygon to indicate the object's boundary. 
For example, \textit{Meng et al.} proposed a \textit{CNN} and Graph Convolution Network (\textit{GCN}) aggregated network \cite{meng2020regression} to directly regress the vertices' coordinates of the OC and OD boundaries.
It is well established that boundary knowledge is essential in obtaining spatial features in segmentation tasks. With regard to biomedical image segmentation, the boundary accuracy is often more critical than that of the regional pixel-wise coverage \cite{meng2020cnn}. 
In this work, we explicitly exploit the boundary information to fuse the boundary features into the proposed graph convolution, proposing a boundary-aware graph convolution.
Our experimental results prove that the proposed boundary-aware graph convolution can improve the region segmentation performance. 
Along the same line, recent methods, such as  \cite{zhang2019net,fang2019selective,wang2021boundary,murugesan2019psi,Meng_2021_TMI}, considered object's boundary segmentation while tackling object region segmentation tasks. However, they treated it as a multi-task learning problem, leaving the model to learn a robust backbone. Differently, we explicitly fuse the boundary information to improve the classic graph convolution's performance on the segmentation task.

This work proposes a novel \textit{GCN} that reasons the boundary-aware region-wise correlations of different classes through learning graph representations. 
Specifically, with an input image, multi-level features were learned through a Backbone Module. Further, we constructed two feature aggregation modules upon shallow- and deep-level backbone features respectively for boundary and region characteristics learning, namely Boundary Segmentation Module \textit{(BSM)} and Region Segmentation Module \textit{(RSM)}. The outputs of \textit{BSM} and \textit{RSM} were fed into the proposed graph reasoning module \textit{GRM} for long range feature reasoning. In detail, the region-wise output ($\textit{R}_{S}$) from \textit{RSM} and the boundary-wise output ($\textit{B}_{S}$) from \textit{BSM} were embedded into graph representation in terms of vertex embeddings.
Within the \textit{GRM}, we proposed a new Boundary-aware Input-dependent Graph Convolution (\textit{BI-GConv}). Specifically, the input-dependent adjacency matrix was estimated from the input features of vertex embeddings that were refined by spatial and channel attentions. 
Additionally, the object's boundary features were fused into the constructed \textit{Laplacian} matrix $\tilde{L}$, enhancing the characteristics along the boundaries to emphasize the boundary-aware correlations across different regions. 
Our experimental results show that the proposed framework makes a large improvement over state-of-the-art approaches. 

\begin{figure*}[t]
\centering
\includegraphics[width=13cm]{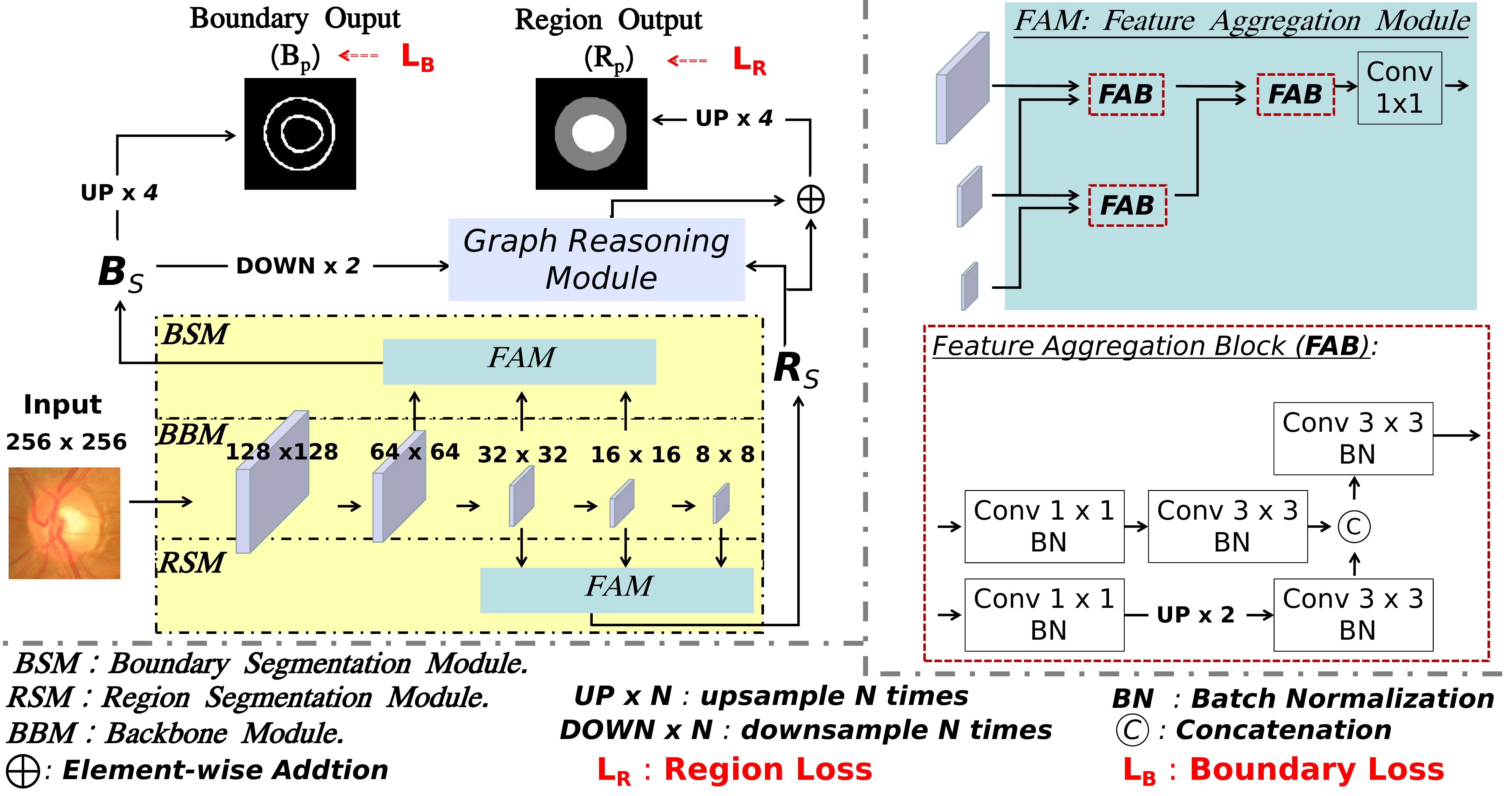}
\caption{Overview of the proposed graph-based segmentation network (left), Feature Aggregation Module (right). Upsample and downsample layers are implemented through learnable bilinear interpolation.  
}
\label{network}
\end{figure*}

\section{Method}
\begin{figure*}[h]
\centering
\includegraphics[width=12cm]{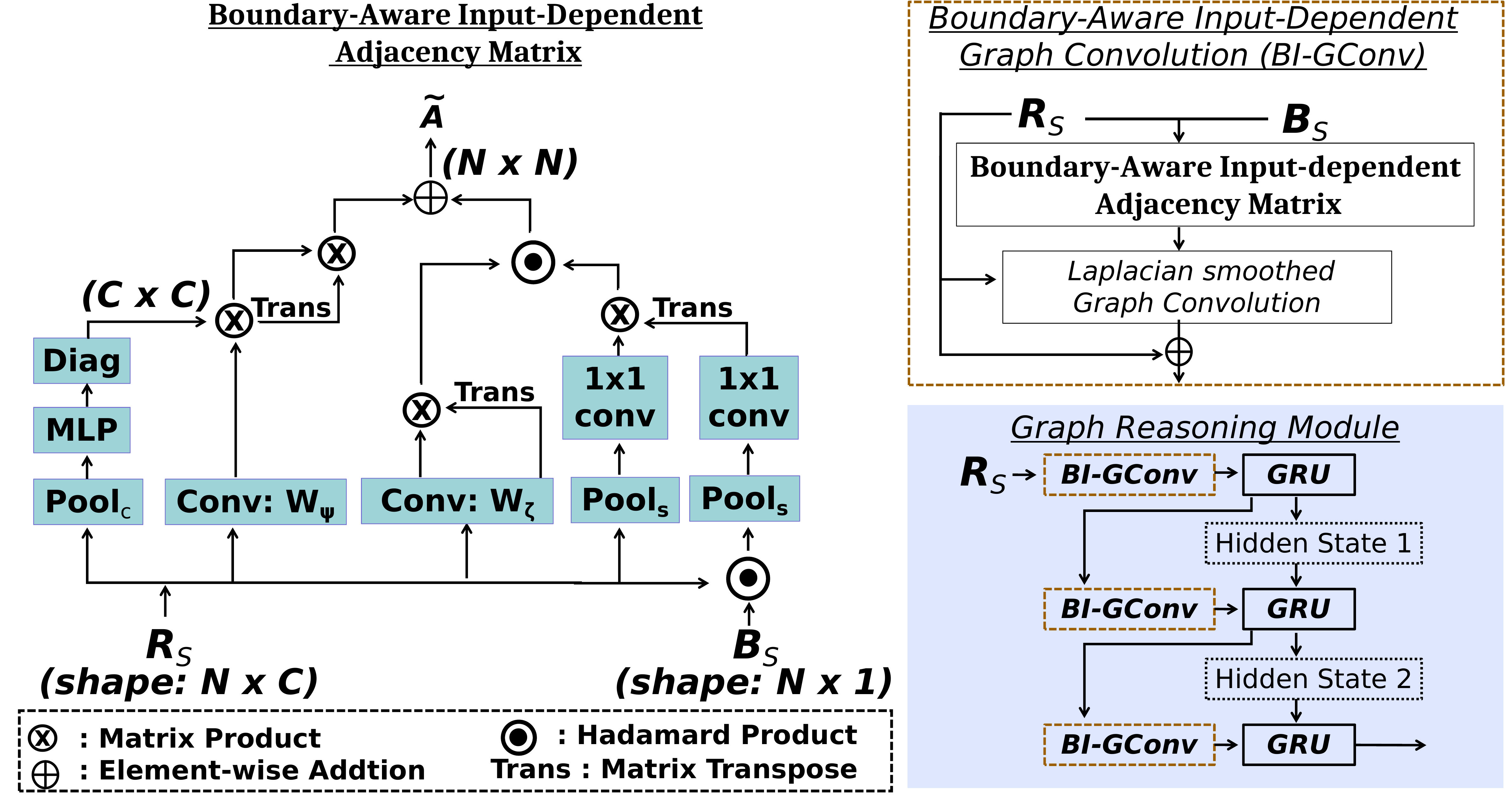}
\caption{Overview of the newly proposed Boundary-Aware Input-Dependent Adjacency Matrix, \textit{BI-GConv} and \textit{GRM}.}

\label{GCN}
\end{figure*}

There are four major components in the proposed model: a backbone module, a region segmentation module (\textit{RSM}), a boundary segmentation module (\textit{BSM}), and a graph reasoning module (\textit{GRM}). Fig. \ref{network} provides an overview of the model architecture and Fig. \ref{GCN} highlights the proposed \textit{GRM} and \textit{BI-GConv}. 
Given an input image, we extract its multi-level features through a backbone module. Following the previous PraNet \cite{fan2020pranet}, we adopt the truncated Res2Net \cite{gao2019res2net} as backbone module due to its superior ability to extract discriminative features in segmentation tasks. 

\noindent\textbf{Feature Aggregation.}
For the \textit{RSM} and \textit{BSM}, we applied two feature aggregation modules to different levels of features of the backbone to extract full spectrum of semantic and spatial characteristics. 
As observed by previous studies \cite{zhou2018unet++,yu2018deep}, deep- and shallow-layer features from different levels complement each other, where the deep-layer features contain rich semantic region information while the shallow-layer features preserve the sufficient spatial boundary information. Thus, we applied feature aggregation module on relative deep-level and low-level features for \textit{BSM} and \textit{RSM} respectively. In detail, the feature aggregation module contains several feature aggregation blocks to fuse low-to-high level features hierarchically and iteratively. Notably, in the feature aggregation block, two $1 \times 1$ convolutions are applied on two resolution's features respectively to reduce and maintain the same channel size. After feature aggregation module, the output \textit{$B_{s}$} and \textit{$R_{s}$} have sizes of \textit{64 $\times$ 64 $\times$ 1} and \textit{32 $\times$ 32 $\times$ 2}, respectively. We downsampled \textit{$B_{s}$} by 2 and fed it into the \textit{GRM} together with \textit{$R_{s}$}. Downsample is for maintaining the same feature map size and reducing the computational overhead. 

\noindent\textbf{Classic Graph Convolution.}
We first revisit the classic graph convolution. Given a graph \textit{G = (V, E)},  normalized \textit{Laplacian} matrix is \(L = I - D ^{ - \frac{1}{2}} A D^{ - \frac{1}{2}}\), where \( I \) is the identity matrix, $A$ is the adjacency matrix, and \(D\) is a diagonal matrix that represents the degree of each vertex in \(V\), such that \(D_{ii} = \sum_{j} A_{i,j}\). 
The \textit{Laplacian} of the graph is a symmetric and positive semi-definite matrix, so \textit{L} can be diagonalized by the Fourier basis $U\in{\mathbb{R}^{N \times N}}$, such that $L = U \Lambda U^{T}$. Thus, the spectral graph convolution of \textit{i} and \textit{j} can be defined as \(i*j = U((U ^{T} i)\) \(\odot\) \((U ^{T} j))\) in the Fourier space. The columns of \(U\) are the orthogonal eigenvectors \(U = [u_1,...,u_n]\), and \(\Lambda = diag([\lambda_{1},...,\lambda_{n}]) \in\mathbb{R} ^{N \times N}\) is a diagonal matrix with non-negative eigenvalues. Since \(U\) is not a sparse matrix, this operation is computationally expensive.
\textit{Defferrard et al.} \cite{defferrard2016convolutional} proposed that the convolution operation on a graph can be defined by formulating spectral filtering with a kernel \(g_{\theta}\) using a recursive Chebyshev polynomial in Fourier space. The filter \(g_{\theta}\) is parameterized as a Chebyshev polynomial expansion of order \(K\), such that $g_{\theta}(L) = \sum_{k} \theta_{k} T_{k} (\hat{L})$, where \(\theta \in \mathbb{R} ^{K}\) is a vector of Chebyshev coefficients, and \(\hat{L} = 2L/\lambda_{max} - I_{N}\) represents the rescaled \textit{Laplacian}. \(T_{k} \in\mathbb{R} ^{N \times N}\) is the Chebyshev polynomial of order \(K\). In \cite{kipf2016semi}, \textit{Kipf et al.} further simplified the graph convolution as $g_{\theta} = \theta(\hat{D} ^{ - \frac{1}{2}} \hat{A} \hat{D}^{ - \frac{1}{2}})$, where $\hat{A} = A + I$, $\hat{D}_{ii} = \sum_{j}\hat{A}_{ij}$, and $\theta$ is the only Chebyshev coefficient left. The corresponding graph \textit{Laplacian} adjacency matrix $\hat{A}$ is hand-crafted, which leads the model to learn a specific long range context pattern rather than the input-related one \cite{li2020spatial}. Thus, we refer to the classic graph convolution as input-independent graph convolution.

\noindent\textbf{Boundary-Aware Input-dependent Graph Convolution (\textit{BI-GConv}).}
Given the input region features $R_{s} \in \mathbb{R}^{N \times C} $, where $C$ is the channel size; $N = H \times W$ is the number of spatial locations of input feature, which is referred to as the number of vertexes. Firstly, we construct the input-dependent adjacency matrix. In detail, we implement two diagonal matrices ($\tilde{\Lambda}^{c}$ and $\tilde{\Lambda}^{s}$) to perform channel-wise attention on the dot-product distance of input vertex embeddings, and measure spatial-wise weighted relations among different vertices, respectively. For example, $\tilde{\Lambda}^{c}(R_{s}) \in \mathbb{R}^{C \times C}$ is the diagonal matrix that contains channel-wise attention on the dot-product distance of the input vertex embeddings; $\tilde{\Lambda}^{s}(R_{s}) \in \mathbb{R}^{N \times N}$ is the spatial-wise weighted matrix, measuring the relationships among different vertices.  
\begin{equation}
    \tilde{\Lambda}^{c}(R_{s}) = diag \Big(MLP\big(Pool_{c}(R_{s})\big)\Big),
\end{equation}
where $Pool_{c}(\cdot)$ denotes the global max pooling for each vertex embedding; $MLP(\cdot)$ is a multi-layer perceptron with one hidden layer. On the other hand,
\begin{equation}
    \tilde{\Lambda}^{s}(R_{s}) = diag \Big(Conv\big(Pool_{s}(R_{s})\big)\Big),
\end{equation}
where $Pool_{s}(\cdot)$ represents the global max pooling for each position in the vertex embedding along the channel axis; $Conv(\cdot)$ is a $1\times1$ convolution layer. The data-dependent adjacency matrix $\Bar{A}$ is given by spatial and channel attention-enhanced input vertex embeddings.
We initialize the input-dependent adjacency matrix $\Bar{A}$ as:
\begin{equation}
    \Bar{A} = \psi(R_{s}, W_{\psi}) \cdot \tilde{\Lambda}^{c}(R_{s}) \cdot \psi(R_{s}, W_{\psi})^{T} + \phi(R_{s}, W_{\phi}) \cdot \phi(R_{s}, W_{\phi})^{T} \odot \tilde{\Lambda}^{s}(R_{s})   ,
\end{equation}
where $\cdot$ represents matrix product; $\odot$ denotes Hadamard product; 
$\psi(R_{s}, W_{\psi}) \in \mathbb{R}^{N \times C}$ and $\phi(R_{s}, W_{\phi}) \in \mathbb{R}^{N \times C}$ are both linear embeddings (1 $\times$ 1 convolution); $W_{\psi}$ and $W_{\phi}$ are learnable parameters.

Secondly, we fuse the boundary information through integrating the predicted boundary function map $\textit{\textbf{B}} {s} \in \mathbb{R}^{N \times 1}$ into the built \textit{Laplacian} matrix $\tilde{L}$, which allows us to leverage the spatial hint of the object's boundary.
Specifically, we fuse it into the spatial-wise weighted matrix $\tilde{\Lambda}^{s}(R_{s})$. The boundary-aware spatial weighted matrix $\tilde{\Lambda}^{s}_{b}(R_{s},B_{s})$ is given as follows: 
\begin{equation}
    \tilde{\Lambda}^{s}_{b}(R_{s},B_{s}) = Conv\Big(Pool_{s}(R_{s})\Big) \cdot  \bigg(Conv\Big(Pool_{s}(R_{s} \odot B_{s})\Big)\bigg)^{T}
\end{equation}
where $\odot$ is the broadcasting Hadamard product along channels; $Conv(\cdot)$ is a $1\times1$ convolution layer.
In this way, the features of boundary pixels are emphasized by assigning larger weights.
As this is the case, the proposed boundary-aware graph convolution can take the spatial characteristics of the boundary into account when reasoning the correlations between different regions. Then, the boundary-aware input-dependent adjacency matrix $\tilde{A}$ will be given as:
\begin{equation}
    \tilde{A} = \psi(R_{s}, W_{\psi}) \cdot \tilde{\Lambda}^{c}(R_{s}) \cdot \psi(R_{s}, W_{\psi})^{T} + \zeta(R_{s}, W_{\zeta}) \cdot \zeta(R_{s}, W_{\zeta})^{T} \odot \tilde{\Lambda}^{s}_{b}(R_{s},B_{s})   ,
\end{equation}
where $\zeta(R_{s}, W_{\zeta}) \in \mathbb{R}^{N \times C}$ is 1 $\times$ 1 convolution; $W_{\zeta}$ is learnable parameter.  Fig. \ref{GCN} shows a detailed demonstration of $\tilde{A}$.
With the constructed $\tilde{A}$, the normalized \textit{Laplacian} matrix is given as $\tilde{L} = I - \tilde{D} ^{ - \frac{1}{2}} \tilde{A} \tilde{D}^{ - \frac{1}{2}}$, where \( I \) is the identity matrix, $\tilde{D}$ is a diagonal matrix that represents the degree of each vertex, such that \(\tilde{D}_{ii} = \sum_{j} \tilde{A}_{i,j}\). 
Inspired by \cite{li2020spatial}, in order to reduce the complexity ($\mathcal{O}(N^{2})$) of calculating \textit{Laplacian} matrix $\tilde{L}$, we calculate the degree matrix $\Tilde{D}$ as:
\begin{equation}
\tilde{D} = diag\bigg(\psi\Big(\tilde{\Lambda}^{c}(\psi^{T} \cdot \Vec{1})\Big)\bigg) + diag\Big(\zeta(\zeta^{T} \cdot \Vec{1})\Big) \odot \tilde{\Lambda}^{s}_{b}
\end{equation}
where $\Vec{1}$ denotes all-one vector in $\mathbb{R}^{N}$, and the calculation in the inner bracket is executed first. By doing this, we can override the quadratic order of computation, thus reduce the computational overhead. Given computed \textit{$\tilde{L}$}, with $R_{s}$ as the input vertex embeddings, we formulate the single-layer \textit{BI-GConv} as :
\begin{equation}
    Y= \sigma (\tilde{L} \cdot R_{s} \cdot W_{G} ) + R_{s},
\end{equation}
where $W_{G}\in \mathbb{R}^{C \times C}$ denotes the trainable weights of the \textit{BI-GConv}; $\sigma$ is the ReLu activation function; \textit{Y} is the output vertex features. Moreover, we add a residual connection to reserve the features of input vertices. 

\noindent\textbf{Graph Reasoning Module.}
We build the \textit{GRM} with the chained \textit{BI-GConv} layers in order to focus on the long-term dependency between chained graph layers. 
Fig. \ref{GCN} shows an overview of the proposed \textit{GRM} with three \textit{BI-GConv} layers of Gated Recurrent Unit (\textit{GRU}) \cite{BallasYPC15} connection, which gains the best performance. Please note, these layers can also be connected by \textit{Residual} \cite{he2016deep} connections; comprehensive comparison results of these two connection manners can be found in the supplementary material.

\section{Experiments}
\subsection{Datasets}

We use a retinal and a colorectal dataset, each of them is merged from five different datasets of its kind. The details of boundary ground truth generation can be found in the supplementary material.

\noindent\textbf{Fundus OD and OC images:} 2,068 images from five datasets (Refuge \cite{orlando2020refuge}, Drishti-GS \cite{sivaswamy2014drishti}, ORIGA \cite{zhang2010origa}, RIGA \cite{almazroa2018retinal}, RIM-ONE \cite{fumero2011rim}) are pooled together. 613 images are randomly selected for test and the remaining 1,455 are for training and validation. We follow \cite{meng2020regression} to crop a subimage of 256 $\times$ 256 pixels centered at the OD center for the subsequent analysis. The training and test data come from the same data source (\textbf{seen data}), OD and OC segmentation is used for evaluate the model's learning ability.

\noindent\textbf{Colonoscopy polyp images:} 2,247 colonoscopy images from five datasets (ETIS \cite{silva2014toward}, CVC-ClinicDB \cite{bernal2015wm}, CVC-ColonDB \cite{tajbakhsh2015automated}, EndoScene-CVC300 \cite{vazquez2017benchmark}, and Kvasir \cite{jha2020kvasir}). All the images are resized to 256 $\times$ 256 pixels. As suggested by \cite{fan2020pranet}, \textit{i.e.}, 1,450 images from Kvasir \cite{jha2020kvasir} and CVC-ClinicDB \cite{bernal2015wm} are selected as the training and validation datasets and the other 635 images from ETIS \cite{silva2014toward}, CVC-ColonDB \cite{tajbakhsh2015automated}, EndoScene-CVC300 \cite{vazquez2017benchmark} are pooled together for independent testing (\textbf{unseen data}). By doing this, the training and test data are from different data source so as to evaluate the model's generalization capability.

\subsection{Implement Details and Evaluation Metrics}
\noindent\textbf{Implement Details} To augment the dataset, we randomly rotate and horizontally flip the training dataset with the probability of 0.3. The rotation ranges from $-30$ to $30$ degree. 10\% of the training dataset are randomly selected as the validation dataset. The network is trained end-to-end by Adam optimizer \cite{kingma2014adam} for 50 thousands iterations, with a learning rate of 6e-3 and decayed to 0.36e-3 after 30 thousand iterations. The batch size is set at 48. All the training processes are performed on a server with four TESLA V100, and all the test experiments are conducted on a local workstation with \textit{Intel(R) Xeon(R) W-2104 CPU} and \textit{Geforce RTX 2080Ti GPU}. \textbf{Code is available at:} \url{https://github.com/smallmax00/BI-GConv}

\noindent\textbf{Loss Function.} 
The overall loss function is defined as: $L_{total} = L_{R} + \alpha \cdot L_{B}$, 
where $\alpha$ balances the losses between region and boundary predictions; a detailed ablation study on $\alpha$ is presented in the supplementary material. We adopt standard Dice Loss \cite{milletari2016v} for the region ($L_{R}$) and boundary ($L_{B}$) segmentation supervision. 

\noindent\textbf{Evaluation Metrics.} Typical region segmentation metrics such as Dice similarity score (\textit{\textbf{Dice}}) and balanced accuracy (\textit{\textbf{B-Acc}}) are applied, where \textit{B-Acc} is the mean value of \textit{Sensitivity} and \textit{Specificity}. Boundary Intersection-over-Union (\textit{\textbf{BIoU}}) \cite{cheng2021boundary} is used as the boundary segmentation measurement. 
Details of the utilization of \textit{BIoU} can be found in the supplementary material. To show statistical significance, 95\% confidence interval is generated by using 2000 sample bootstrapping for all the evaluation metrics. 

\subsection{Results}
\begin{figure*}[h]
\centering
\includegraphics[width=10cm]{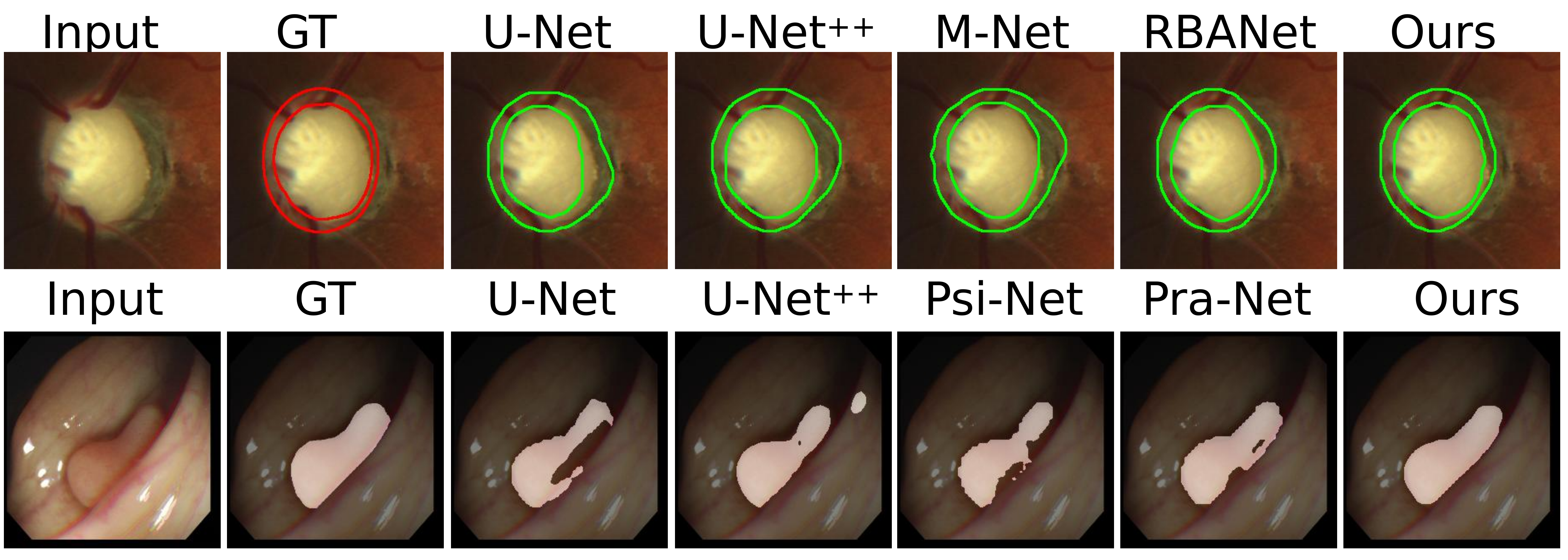}
\caption{Visualization of OC and OD segmentation (first row), where the boundary of region segmentation predictions are extracted and plotted on the input image for better visualization. Second row shows the colonoscopy polyp segmentation comparisons.
Our proposed method consistently produces segmentation results closest to the ground truth.}
\label{qua_results}
\end{figure*}

We show qualitative (Fig. \ref{qua_results}) and quantitative results (Tab. \ref{quantitative}, Tab. \ref{complexity}) of the polyps, and OC and OD segmentation tasks. More qualitative results can be found in the supplementary materials.
We compared the proposed method with two classic models (\textit{i.e.} \textit{U-Net} \cite{ronneberger2015u},\textit{U-Net++} \cite{zhou2018unet++}) and five cutting-edge models, such as \textit{M-Net} \cite{fu2018joint}, \textit{RBA-Net} \cite{meng2020regression}, \textit{PsiNet} \cite{murugesan2019psi}, \textit{PraNet} \cite{fan2020pranet} and \textit{ACSNet} \cite{zhang2020adaptive} for OC and OD and polyps segmentation. Notably, we sampled 360 vertices for \textit{RBA-Net} \cite{meng2020regression} to construct a smooth boundary. \textit{M-Net} \cite{fu2018joint} is specially designed for OC and OD segmentation task. 


\noindent\textbf{Learning Ability.}
We conduct experiments to validate our model’s learning ability on \textit{seen} data (\textit{i.e.} OC and OD segmentation datasets).
As shown in Tab. \ref{quantitative}, we achieved 88.8 \% and 97.7 \% \textit{Dice} scores on OC and OD segmentation respectively, which outperforms region segmentation based approaches \textit{U-Net} \cite{ronneberger2015u}, \textit{U-Net++} \cite{zhou2018unet++}, \textit{PraNet} \cite{fan2020pranet} and \textit{M-Net} \cite{fu2018joint} by 3.5 \%, 3.1 \%, 2.9 \% and 1.9 \% respectively, boundary regression approach \textit{RBA-Net} \cite{meng2020regression} by 1.9 \%, and region-boundary hybrid approach \textit{PsiNet} \cite{murugesan2019psi} by 2.8 \%.

\noindent\textbf{Generalization Capability.}
We conduct experiments to test the model’s generalizability on \textit{unseen} data (\textit{i.e.} polyps segmentation datasets). The three unseen datasets each have their own set of challenges and characteristics. For example, CVC-ColonDB \cite{tajbakhsh2015automated} is a small-scale database comprised of 15 short colonoscopy sequences.
ETIS \cite{silva2014toward} is an early established dataset for colorectal cancer early diagnosis. 
EndoScene-CVC300 \cite{vazquez2017benchmark} is a re-annotated benchmark with associated polyp and background (mucosa and lumen).
In Tab. \ref{quantitative}, our model achieves 72.7 \% \textit{Dice} score, which outperforms the cutting-edge polyps segmentation methods \textit{PraNet} \cite{fan2020pranet} and \textit{ACSNet} \cite{zhang2020adaptive} by 3.1 \% and 5.2 \%, respectively.

\begin{table*}[t]
\begin{center}
\caption{Quantitative segmentation results of OC and OD and polyps on respective testing datasets. The performance is reported as \textit{Dice} (\%), \textit{B-Acc} (\%), \textit{BIoU} (\%). 95 \% confidence intervals are presented in the brackets, respectively.
}
\label{quantitative}
\centering
\scalebox{0.6}
{
\begin{tabular}{c||ccc || ccc || ccc}
\hline
\multirow{2}{*}{Methods} & \multicolumn{3}{c||}{OC} & \multicolumn{3}{c||}{OD} & \multicolumn{3}{c}{Polyps} \\ \cline{2-10} 
 & \textit{Dice (\%)\textuparrow}  & \textit{B-Acc (\%)\textuparrow} & \textit{BIoU (\%)\textuparrow}
 & \textit{Dice (\%)\textuparrow}  & \textit{B-Acc (\%)\textuparrow} & \textit{BIoU (\%)\textuparrow}
 & \textit{Dice (\%)\textuparrow}  & \textit{B-Acc (\%)\textuparrow} & \textit{BIoU (\%)\textuparrow} \\ \hline
\begin{tabular}[c]{@{}c@{}}\textit{U-Net} \cite{ronneberger2015u} \\ \end{tabular}
&\begin{tabular}[c]{@{}c@{}}85.3\\ (84.0, 85.9)\end{tabular}  
&\begin{tabular}[c]{@{}c@{}}88.1\\ (86.8, 88.9)\end{tabular}  
&\begin{tabular}[c]{@{}c@{}}80.1\\ (78.7, 81.5)\end{tabular}

&\begin{tabular}[c]{@{}c@{}}95.0\\ (94.0, 96.2)\end{tabular}  
&\begin{tabular}[c]{@{}c@{}}96.0\\ (95.1, 96.7)\end{tabular}  
&\begin{tabular}[c]{@{}c@{}}86.2\\ (85.1, 87.3)\end{tabular}

&\begin{tabular}[c]{@{}c@{}}67.7\\ (63.6, 70.1)\end{tabular}
&\begin{tabular}[c]{@{}c@{}}80.7\\ (77.0, 84.9)\end{tabular}   
&\begin{tabular}[c]{@{}c@{}}61.0\\ (58.6, 64.1)\end{tabular} \\ 

\begin{tabular}[c]{@{}c@{}}\textit{U-Net++} \cite{zhou2018unet++} \\ \end{tabular}
&\begin{tabular}[c]{@{}c@{}}86.0\\ (84.7, 87.6)\end{tabular} 
&\begin{tabular}[c]{@{}c@{}}89.6\\ (88.3, 89.9)\end{tabular}  
&\begin{tabular}[c]{@{}c@{}}81.4\\ (79.5, 83.0)\end{tabular}

&\begin{tabular}[c]{@{}c@{}}95.0\\ (93.9, 96.1)\end{tabular} 
&\begin{tabular}[c]{@{}c@{}}95.9\\ (95.0, 96.5)\end{tabular}  
&\begin{tabular}[c]{@{}c@{}}87.0\\ (86.3, 87.8)\end{tabular}

&\begin{tabular}[c]{@{}c@{}}67.6\\ (64.1, 71.0)\end{tabular}
&\begin{tabular}[c]{@{}c@{}}81.6\\ (76.4, 85.5)\end{tabular}   
&\begin{tabular}[c]{@{}c@{}}62.1\\ (59.7, 64.7)\end{tabular}\\ 

\begin{tabular}[c]{@{}c@{}}\textit{M-Net}  \cite{fu2018joint} \\ \end{tabular}
&\begin{tabular}[c]{@{}c@{}}86.9\\ (86.1, 87.2)\end{tabular} 
&\begin{tabular}[c]{@{}c@{}}91.7\\ (91.3, 92.4)\end{tabular}  
&\begin{tabular}[c]{@{}c@{}}82.9\\ (79.5, 84.1)\end{tabular}

&\begin{tabular}[c]{@{}c@{}}96.2\\ (95.8, 96.6)\end{tabular} 
&\begin{tabular}[c]{@{}c@{}}97.2\\ (96.3, 97.9)\end{tabular}  
&\begin{tabular}[c]{@{}c@{}}88.1\\ (87.0, 88.7)\end{tabular}
&\begin{tabular}[c]{@{}c@{}}-\\ \end{tabular}
&\begin{tabular}[c]{@{}c@{}}-\\ \end{tabular}   
&\begin{tabular}[c]{@{}c@{}}-\\ \end{tabular}\\ 

\begin{tabular}[c]{@{}c@{}}\textit{RBA-Net} \cite{meng2020regression} \\ \end{tabular}
&\begin{tabular}[c]{@{}c@{}}87.1\\ (86.2, 88.0)\end{tabular}
&\begin{tabular}[c]{@{}c@{}}92.5\\ (91.1, 92.9)\end{tabular}  
&\begin{tabular}[c]{@{}c@{}}82.8\\ (81.6, 83.9)\end{tabular}

&\begin{tabular}[c]{@{}c@{}}96.1\\ (95.5, 96.7)\end{tabular}
&\begin{tabular}[c]{@{}c@{}}97.5\\ (96.8, 98.2)\end{tabular}  
&\begin{tabular}[c]{@{}c@{}}88.9\\ (87.9, 89.4)\end{tabular}

&\begin{tabular}[c]{@{}c@{}}69.5\\ (67.2, 71.1)\end{tabular}
&\begin{tabular}[c]{@{}c@{}}84.1\\ (83.6, 84.7)\end{tabular}   
&\begin{tabular}[c]{@{}c@{}}63.2\\ (60.8, 65.1)\end{tabular}\\

\begin{tabular}[c]{@{}c@{}}\textit{Psi-Net} \cite{murugesan2019psi} \\ \end{tabular}
&\begin{tabular}[c]{@{}c@{}}85.7\\ (84.3, 86.4)\end{tabular}
&\begin{tabular}[c]{@{}c@{}}88.1\\ (87.5, 88.6)\end{tabular}  
&\begin{tabular}[c]{@{}c@{}}80.1\\ (80.7, 81.0)\end{tabular}

&\begin{tabular}[c]{@{}c@{}}95.8\\ (94.5, 96.1)\end{tabular}
&\begin{tabular}[c]{@{}c@{}}96.7\\ (95.5, 97.0)\end{tabular}  
&\begin{tabular}[c]{@{}c@{}}87.9\\ (86.4, 88.5)\end{tabular}

&\begin{tabular}[c]{@{}c@{}}67.8\\ (64.7, 69.9)\end{tabular}
&\begin{tabular}[c]{@{}c@{}}84.5\\ (81.1, 87.2)\end{tabular}   
&\begin{tabular}[c]{@{}c@{}}61.1\\ (59.7, 63.2)\end{tabular}\\

\begin{tabular}[c]{@{}c@{}}\textit{ACSNet} \cite{zhang2020adaptive} \\ \end{tabular}
&\begin{tabular}[c]{@{}c@{}}85.5 \\ (84.1, 86.7)\end{tabular}
&\begin{tabular}[c]{@{}c@{}}88.8 \\ (88.1, 89.2)\end{tabular}  
&\begin{tabular}[c]{@{}c@{}}81.1 \\ (79.4 , 82.6)\end{tabular}

&\begin{tabular}[c]{@{}c@{}}95.9 \\ (94.5, 97.1)\end{tabular}
&\begin{tabular}[c]{@{}c@{}}96.4 \\ (96.0, 96.7)\end{tabular}  
&\begin{tabular}[c]{@{}c@{}}87.0 \\ (85.9, 88.3)\end{tabular}

&\begin{tabular}[c]{@{}c@{}}69.1 \\ (66.7, 71.4)\end{tabular}
&\begin{tabular}[c]{@{}c@{}}83.6 \\ (79.7, 85.5)\end{tabular}   
&\begin{tabular}[c]{@{}c@{}}65.2 \\ (62.5, 67.7)\end{tabular}\\ 

\begin{tabular}[c]{@{}c@{}}\textit{PraNet} \cite{fan2020pranet} \\ \end{tabular}
&\begin{tabular}[c]{@{}c@{}} 85.8 \\ (84.5, 86.3) \end{tabular}  
&\begin{tabular}[c]{@{}c@{}} 88.9 \\ (88.1, 89.2) \end{tabular}  
&\begin{tabular}[c]{@{}c@{}} 81.5 \\ (80.1, 82.7) \end{tabular}
&\begin{tabular}[c]{@{}c@{}} 95.5 \\ (93.7, 97.3) \end{tabular}  
&\begin{tabular}[c]{@{}c@{}} 97.1 \\ (96.2, 97.9) \end{tabular}  
&\begin{tabular}[c]{@{}c@{}} 86.8 \\ (85.1, 87.7) \end{tabular}

&\begin{tabular}[c]{@{}c@{}}70.5\\ (68.6, 71.7)\end{tabular}
&\begin{tabular}[c]{@{}c@{}}85.6\\ (83.8, 86.9)\end{tabular}   
&\begin{tabular}[c]{@{}c@{}}64.0\\ (61.6, 66.9)\end{tabular}\\

\hline \hline

\begin{tabular}[c]{@{}c@{}}\textit{Ours}  \\ \end{tabular}
&\begin{tabular}[c]{@{}c@{}}\textbf{88.8}\\ (88.1, 89.4)\end{tabular}
&\begin{tabular}[c]{@{}c@{}}\textbf{94.7}\\ (94.2, 95.1)\end{tabular}  
&\begin{tabular}[c]{@{}c@{}}\textbf{85.1}\\ (83.3, 86.8)\end{tabular}
&\begin{tabular}[c]{@{}c@{}}\textbf{97.7}\\ (97.6, 97.8)\end{tabular}
&\begin{tabular}[c]{@{}c@{}}\textbf{98.4}\\ (98.4, 98.5)\end{tabular}  
&\begin{tabular}[c]{@{}c@{}}\textbf{91.1}\\ (90.2, 92.0)\end{tabular}

&\begin{tabular}[c]{@{}c@{}}\textbf{72.7}\\ (70.1, 75.4)\end{tabular}
&\begin{tabular}[c]{@{}c@{}}\textbf{87.5}\\ (86.2, 88.8)\end{tabular}  
&\begin{tabular}[c]{@{}c@{}}\textbf{67.4}\\ (65.7, 70.1)\end{tabular} \\ \hline
\end{tabular}
}
\end{center}
\end{table*}

\noindent\textbf{Computational Efficiency.}
Tab. \ref{complexity} presents the number of parameters (\textit{M}) and floating-point operations (\textit{FLOPs}) of the compared models. \textit{Ours} and \textit{PraNet} \cite{fan2020pranet} adopt the same backbone network, thus have similar model size (\textit{Params}). However, our model iteratively and hierarchically fuse and aggregate deep- and shallow-layers features, which consume more computations. A trade-off between segmentation accuracy and computational efficiency can be determined by different number of chained \textit{BI-GConv} layers. For example, \textit{Ours} contains 26.6 \textit{M} parameters when there is one layer of \textit{BI-GConv} in the proposed \textit{GRM}. 

\begin{table}[]
\centering
\caption{Number of parameters and \textit{FLOPs} on a 256 $\times$ 256 input image.}
\label{complexity}
\scalebox{0.7}
{
\begin{tabular}{l|lllllll|l}
\hline
 & \textit{U-net} \cite{ronneberger2015u} & \textit{U-Net++} \cite{zhou2018unet++} & \textit{M-net} \cite{fu2018joint} & \textit{RBA-Net} \cite{meng2020regression} & \textit{Psi-Net} \cite{murugesan2019psi} & \textit{ACSnet} \cite{zhang2020adaptive} & \textit{PraNet} \cite{fan2020pranet} & \textit{Ours} \\ \hline
\textit{Params} (\textit{M}) &17.3  &36.6  &10.3 &34.3  &7.8  &29.5  &30.5  & 32.9  \\ \hline
\textit{FLOPs} (\textit{G}) &40.2  &138.6  &17.5 &130.3  &32.6  &11.5  & 6.9 & 12.0  \\ \hline
\end{tabular}
}
\end{table}

\subsection{Ablation Study}\label{sec:ablation}
We conducted extensive ablation studies, and all the results consolidated our model's effectiveness. For example, the ablation results on different graph convolutions and model components are shown in Tab. \ref{ablation1} and Tab. \ref{ablation2}. In the supplementary material, we report the ablation study experimental results related to the \textit{GRM}, such as number (\textit{N}) of \textit{BI-GConv} layers, different multi-layers connection manners (\textit{GRU} \cite{BallasYPC15} or \textit{Residual} \cite{he2016deep}).

\begin{table}[h]
\caption{
Ablation study on different graph convolutions. The performance is reported as \textit{Dice (\%)} and \textit{BIoU (\%)}. 95 \% confidence intervals are presented in the brackets, respectively.}
\label{ablation1}
\begin{center}
\centering
\scalebox{0.7}
{
\begin{tabular}{c||cc|cc|cc}
\hline
\multirow{2}{*}{Methods} & \multicolumn{2}{c|}{OC} & \multicolumn{2}{c|}{OD} & \multicolumn{2}{c}{Polyps} \\ \cline{2-7} 
&\textit{Dice (\%)\textuparrow}  & \textit{BIoU (\%)\textuparrow} & \textit{Dice (\%)\textuparrow}  & \textit{BIoU (\%)\textuparrow} & \textit{Dice (\%)\textuparrow} & \textit{BIoU (\%)\textuparrow} \\ \hline
\begin{tabular}[c]{@{}c@{}}\textit{Classic Graph Convolution} \cite{kipf2016semi}  \\ \end{tabular}
&\begin{tabular}[c]{@{}c@{}} 86.5 \\ (85.5, 87.3) \end{tabular}  
&\begin{tabular}[c]{@{}c@{}} 81.1 \\ (79.9, 82.2) \end{tabular}  
&\begin{tabular}[c]{@{}c@{}} 95.5 \\ (95.0, 96.1) \end{tabular}
&\begin{tabular}[c]{@{}c@{}} 86.4 \\ (85.3, 87.3) \end{tabular}  
&\begin{tabular}[c]{@{}c@{}} 69.5 \\ (66.9, 71.6) \end{tabular}  
&\begin{tabular}[c]{@{}c@{}} 63.8 \\ (61.1, 66.0) \end{tabular} \\

\begin{tabular}[c]{@{}c@{}}\textit{Dependent (channel)}  \\ \end{tabular}
&\begin{tabular}[c]{@{}c@{}} 87.3 \\ (86.5, 87.9) \end{tabular}  
&\begin{tabular}[c]{@{}c@{}} 83.0 \\ (81.2, 85.2) \end{tabular}  
&\begin{tabular}[c]{@{}c@{}} 95.9 \\ (95.2, 96.5) \end{tabular}
&\begin{tabular}[c]{@{}c@{}} 86.8 \\ (85.7, 87.4) \end{tabular}  
&\begin{tabular}[c]{@{}c@{}} 69.9 \\ (67.2, 71.9) \end{tabular}  
&\begin{tabular}[c]{@{}c@{}} 64.0 \\ (61.1, 66.7) \end{tabular}\\

\begin{tabular}[c]{@{}c@{}}\textit{Dependent (spatial)}  \\ \end{tabular}
&\begin{tabular}[c]{@{}c@{}} 87.8 \\ (86.9, 88.3) \end{tabular}  
&\begin{tabular}[c]{@{}c@{}} 83.5 \\ (81.9, 85.0) \end{tabular}  
&\begin{tabular}[c]{@{}c@{}} 96.1 \\ (95.6, 96.8) \end{tabular}
&\begin{tabular}[c]{@{}c@{}} 86.9 \\ (85.6, 87.7) \end{tabular}  
&\begin{tabular}[c]{@{}c@{}} 70.2 \\ (67.1, 72.9) \end{tabular}  
&\begin{tabular}[c]{@{}c@{}} 64.2 \\ (62.1, 66.7) \end{tabular}\\

\begin{tabular}[c]{@{}c@{}}\textit{Dependent (channel \& spatial)}  \\ \end{tabular}
&\begin{tabular}[c]{@{}c@{}} 88.1 \\ (87.3, 88.8) \end{tabular}  
&\begin{tabular}[c]{@{}c@{}} 83.9 \\ (81.7, 85.0) \end{tabular}  
&\begin{tabular}[c]{@{}c@{}} 96.5 \\ (96.1, 97.1) \end{tabular}
&\begin{tabular}[c]{@{}c@{}} 87.7 \\ (86.4, 88.6) \end{tabular}  
&\begin{tabular}[c]{@{}c@{}} 70.8 \\ (67.9, 72.9) \end{tabular}  
&\begin{tabular}[c]{@{}c@{}} 64.5 \\ (61.8, 66.5) \end{tabular}\\ \hline

\begin{tabular}[c]{@{}c@{}}w/ \textit{SGR} \cite{liang2018symbolic}  \\ \end{tabular}
&\begin{tabular}[c]{@{}c@{}} 87.5 \\ (86.2, 88.0) \end{tabular}  
&\begin{tabular}[c]{@{}c@{}} 83.3 \\ (81.6, 85.2) \end{tabular}  
&\begin{tabular}[c]{@{}c@{}} 96.4 \\ (96.1, 96.7) \end{tabular}
&\begin{tabular}[c]{@{}c@{}} 87.6 \\ (86.7, 88.5) \end{tabular}  
&\begin{tabular}[c]{@{}c@{}} 70.7 \\ (67.1, 72.9) \end{tabular}  
&\begin{tabular}[c]{@{}c@{}} 64.4 \\ (61.1, 66.7) \end{tabular}\\

\begin{tabular}[c]{@{}c@{}}w/ \textit{DualGCN} \cite{zhangli_dgcn} \\ \end{tabular}
&\begin{tabular}[c]{@{}c@{}} 87.8 \\ (86.9, 88.3) \end{tabular}  
&\begin{tabular}[c]{@{}c@{}} 83.9 \\ (81.7, 85.8) \end{tabular}  
&\begin{tabular}[c]{@{}c@{}} 96.8 \\ (96.4, 97.0) \end{tabular}
&\begin{tabular}[c]{@{}c@{}} 88.0 \\ (87.2, 88.8) \end{tabular}  
&\begin{tabular}[c]{@{}c@{}} 70.9 \\ (67.2, 73.1) \end{tabular}  
&\begin{tabular}[c]{@{}c@{}} 65.0 \\ (63.1, 67.7) \end{tabular}\\

\begin{tabular}[c]{@{}c@{}}w/ \textit{GloRe} \cite{chen2019graph} \\ \end{tabular}
&\begin{tabular}[c]{@{}c@{}} 87.9 \\ (86.5, 88.2) \end{tabular}  
&\begin{tabular}[c]{@{}c@{}} 84.1 \\ (82.6, 86.0) \end{tabular}  
&\begin{tabular}[c]{@{}c@{}} 96.9 \\ (96.5, 97.4) \end{tabular}
&\begin{tabular}[c]{@{}c@{}} 88.3 \\ (87.6, 88.9) \end{tabular}  
&\begin{tabular}[c]{@{}c@{}} 71.2 \\ (68.1, 73.2) \end{tabular}  
&\begin{tabular}[c]{@{}c@{}} 66.1 \\ (64.3, 67.7) \end{tabular}\\ \hline \hline

\begin{tabular}[c]{@{}c@{}}\textit{Ours}  \\ \textit{(channel \& spatial) + boundary} \end{tabular}
&\begin{tabular}[c]{@{}c@{}}\textbf{88.8}\\ (88.1, 89.4)\end{tabular}
&\begin{tabular}[c]{@{}c@{}}\textbf{85.1}\\ (83.3, 86.8)\end{tabular}
&\begin{tabular}[c]{@{}c@{}}\textbf{97.7}\\ (97.6, 97.8)\end{tabular}
&\begin{tabular}[c]{@{}c@{}}\textbf{91.1}\\ (90.2, 92.0)\end{tabular}
&\begin{tabular}[c]{@{}c@{}}\textbf{72.7}\\ (70.1, 75.4)\end{tabular}
&\begin{tabular}[c]{@{}c@{}}\textbf{67.4}\\ (65.7, 70.1)\end{tabular} \\ \hline

\end{tabular}
}
\end{center}
\end{table}

\noindent\textbf{Graph Convolution.}
In this section, we evaluate the effectiveness of the proposed graph convolution operation. Firstly, we employ the classic graph convolution \cite{kipf2016semi} to reason the correlations between regions. Then we explore the input-dependent graph convolutions in terms of channel attention (\textit{channel}) and spatial attention (\textit{spatial}) mechanism, respectively and simultaneously (\textit{channel and spatial}). 
Additionally, we adopt three more potent \textit{GRMs} to show the superiority of our proposed \textit{BI-GConv}. In detail, we apply the \textit{SGR} \cite{liang2018symbolic}, \textit{DualGCN} \cite{zhangli_dgcn}, and \textit{GloRe} module \cite{chen2019graph} respectively, where the \textit{SGR} module exploits knowledge graph mechanism; \textit{DualGCN} explores the coordinate space and feature space graph convolution; and \textit{GloRe} utilizes projection and re-projection mechanism to reason the semantics between different regions. These methods achieved state-of-the-art performance on different segmentation tasks, however, they all belong to the data-independent \textit{Laplacian} based graph convolution.
Tab. \ref{ablation1} shows that our model achieves more accurate and reliable results than \cite{kipf2016semi} and outperforms the \textit{SGR}, \textit{DualGCN}, and \textit{GloRe} by 1.9 \%, 1.5 \% and 1.3 \% in terms of mean \textit{Dice} on the two test datasets. 

\begin{table}[h]
\caption{
Ablation study on different graph convolutions. The performance is reported as \textit{Dice (\%)} and \textit{BIoU (\%)}. 95\% confidence intervals are presented in the brackets, respectively.}
\label{ablation2}
\begin{center}
\centering
\scalebox{0.7}
{
\begin{tabular}{c||cc|cc|cc}
\hline
\multirow{2}{*}{Methods} & \multicolumn{2}{c|}{OC} & \multicolumn{2}{c|}{OD} & \multicolumn{2}{c}{Polyps} \\ \cline{2-7} 
&\textit{Dice (\%)\textuparrow}  & \textit{BIoU (\%)\textuparrow} & \textit{Dice (\%)\textuparrow}  & \textit{BIoU (\%)\textuparrow} & \textit{Dice (\%)\textuparrow} & \textit{BIoU (\%)\textuparrow} \\ \hline
\begin{tabular}[c]{@{}c@{}}\textit{Baseline} \\ \end{tabular}
&\begin{tabular}[c]{@{}c@{}} 85.1 \\ (84.0, 86.3) \end{tabular}  
&\begin{tabular}[c]{@{}c@{}} 80.0 \\ (78.1, 81.2) \end{tabular}  
&\begin{tabular}[c]{@{}c@{}} 95.5 \\ (95.0, 96.0) \end{tabular}
&\begin{tabular}[c]{@{}c@{}} 86.3 \\ (85.8, 86.7) \end{tabular}  
&\begin{tabular}[c]{@{}c@{}} 69.1 \\ (66.9, 71.5) \end{tabular}  
&\begin{tabular}[c]{@{}c@{}} 63.4 \\ (61.1, 65.7) \end{tabular}\\

\begin{tabular}[c]{@{}c@{}}w/ \textit{BSM} \\ \end{tabular}
&\begin{tabular}[c]{@{}c@{}} 85.8 \\ (84.5, 86.3) \end{tabular}  
&\begin{tabular}[c]{@{}c@{}} 81.5 \\ (80.2, 82.2) \end{tabular}  
&\begin{tabular}[c]{@{}c@{}} 96.1 \\ (95.4, 96.7) \end{tabular}
&\begin{tabular}[c]{@{}c@{}} 87.7 \\ (87.2, 88.3) \end{tabular}  
&\begin{tabular}[c]{@{}c@{}} 70.3 \\ (69.0, 72.1) \end{tabular}  
&\begin{tabular}[c]{@{}c@{}} 64.1 \\ (62.0, 65.7) \end{tabular}\\

\begin{tabular}[c]{@{}c@{}}w/ \textit{BI-GConv} \\ (\textit{N} = 1) \end{tabular}
&\begin{tabular}[c]{@{}c@{}} 87.9 \\ (85.5, 89.3) \end{tabular}  
&\begin{tabular}[c]{@{}c@{}} 83.8 \\ (81.1, 85.2) \end{tabular}  
&\begin{tabular}[c]{@{}c@{}} 97.0 \\ (96.6, 97.3) \end{tabular}
&\begin{tabular}[c]{@{}c@{}} 90.2 \\ (89.8, 90.6) \end{tabular}  
&\begin{tabular}[c]{@{}c@{}} 72.1 \\ (70.0, 74.7) \end{tabular}  
&\begin{tabular}[c]{@{}c@{}} 66.7 \\ (64.1, 68.1) \end{tabular}\\ \hline \hline

\begin{tabular}[c]{@{}c@{}}\textit{Ours}  \\ w/ \textit{BI-GConv} (N =3)\end{tabular}
&\begin{tabular}[c]{@{}c@{}}\textbf{88.8}\\ (88.1, 89.4)\end{tabular}
&\begin{tabular}[c]{@{}c@{}}\textbf{85.1}\\ (83.3, 86.8)\end{tabular}
&\begin{tabular}[c]{@{}c@{}}\textbf{97.7}\\ (97.6, 97.8)\end{tabular}
&\begin{tabular}[c]{@{}c@{}}\textbf{91.1}\\ (90.2, 92.0)\end{tabular}
&\begin{tabular}[c]{@{}c@{}}\textbf{72.7}\\ (70.1, 75.4)\end{tabular}
&\begin{tabular}[c]{@{}c@{}}\textbf{67.4}\\ (65.7, 70.1)\end{tabular} \\ \hline


\end{tabular}
}
\end{center}
\end{table}

\noindent\textbf{Model Components.}
We evaluate the effectiveness of each component of our proposed model. Firstly, we remove the \textit{BSM} and the \textit{GRM}; we refer to the rest of the structure as the Baseline model. Then we add \textit{BSM} (w/ \textit{BSM}) and single layer \textit{BI-GConv} (w/ \textit{BI-GConv}, \textit{N} = 1) to evaluate each component's effectiveness gradually.
Note that, in order to eliminate the model size difference and maintain a similar number of parameters (e.g. 32.8 million for our model), we replace the \textit{GRMs} with several feed-forward \textit{CNN} blocks in other ablation models (\textit{e.g.}, \textit{Baseline}, w/ \textit{BSM}, \textit{etc.}) with standard convolution layer of kernel size 3 $\times$ 3, padding 1, followed by a Batch Normalization \cite{ioffe2015batch} layer.
Tab. \ref{ablation2} shows that \textit{BSM} and single layer \textit{BI-GConv} can improve from Baseline model by 1.0 \% and 3.4 \% via the mean \textit{Dice} on the two test datasets.

\section{Conclusion}
In this paper, we present a graph-propagation based framework to tackle biomedical image segmentation problems. We proposed a boundary-aware input-dependent graph convolution (\textit{BI-GConv}) to reason about the boundary-enhanced long range correlations between regions in biomedical image segmentation tasks. Our experiment results show that the proposed \textit{GRM} can efficiently reason about the semantic region features while explicitly considering spatial boundary features on the segmentation tasks of the colonic  polyps and OD and OC. We will extend the proposed model to tackle 3D biomedical image segmentation tasks to learn the graph representation of volumes and object surfaces in the future. 

\section{Acknowledgements}
Y. Meng and H. Zhang thank the China Science IntelliCloud Technology Co., Ltd for the studentships. We thank NVIDIA for the donation of GPU cards. This work was undertaken on Barkla, part of the High Performance Computing facilities at the University of Liverpool, Liverpool, United Kingdom.
\bibliography{egbib}
\end{document}